\documentclass[letterpaper, 10 pt, conference]{ieeeconf}  % Comment this line out if you need a4paper
\IEEEoverridecommandlockouts                              % This command is only needed if 
\usepackage{amssymb,amsmath}
\usepackage{dblfloatfix}  
\usepackage{lineno}
\usepackage[figuresright]{rotating}
\usepackage{epstopdf}
\usepackage{algorithm}
\usepackage{algorithmic}
\usepackage{array, multirow}
\usepackage{xspace}
\usepackage{hyperref}
\usepackage{graphicx}
\usepackage{tabularx}
\usepackage{array}
\usepackage{color}
\usepackage{multicol}
\usepackage{float}
\usepackage{xspace,booktabs}
\usepackage{eso-pic}
\usepackage{pdfpages}
\usepackage{algorithm}
\usepackage[percent]{overpic}
\usepackage{sidecap}
\usepackage{hhline}
\usepackage{makecell}
\usepackage[caption=false,font=footnotesize]{subfig}
\usepackage[figuresright]{rotating}
\usepackage[export]{adjustbox}
\usepackage{stackengine}
\usepackage{tikz}
\usetikzlibrary{shapes,arrows}
\tikzstyle{decision} = [diamond, draw, fill=blue!20, 
    text width=4.5em, text badly centered, node distance=3cm, inner sep=0pt]
\tikzstyle{block} = [rectangle, draw,
    text width=5em, text centered, rounded corners, minimum height=4em]
\tikzstyle{line} = [draw, -latex']
\tikzstyle{cloud} = [draw, ellipse,fill=red!20, node distance=3cm,
    minimum height=2em]

\usepackage{pgfplots}
\pgfplotsset{width=10cm,compat=1.15}
\usepgfplotslibrary{statistics}
\usetikzlibrary{pgfplots.groupplots}

\usepackage[normalem]{ulem}

\title{\LARGE \bf
OHPL: One-shot Hand-eye Policy Learner
}

\author{Changjae Oh, Yik Lung Pang, and Andrea Cavallaro% <-this % stops a space
\thanks{Changjae Oh and Yik Lung Pang equally contributed. The authors are with 
the Centre for Intelligent Sensing,
Queen Mary University of London, E1 4NS, London, United Kingdom. \{{\tt\footnotesize c.oh}, {\tt\footnotesize y.l.pang}, {\tt\footnotesize a.cavallaro}\}{\tt\footnotesize@qmul.ac.uk}.
}
}
\begin{document}

\maketitle

%%%%%%%%%%%%%%%%%%%%%%%%%%%%%%%%%%%%%%%%%%%%%

\begin{abstract}
The control of a robot for manipulation tasks generally relies on object detection and pose estimation. An attractive alternative is to learn control policies directly from raw input data. However, this approach is time-consuming and expensive since learning the policy requires many trials with robot actions in the physical environment. To reduce the training cost, the policy can be learned in simulation with a large set of synthetic images. The limit of this approach is the domain gap between the simulation and the robot workspace. In this paper, we propose to learn a policy for robot reaching movements from a single image captured directly in the robot workspace from a camera placed on the end-effector (a hand-eye camera). The idea behind the proposed policy learner is that view changes seen from the hand-eye camera produced by actions in the robot workspace are analogous to locating a region-of-interest in a single image by performing sequential object localisation. This similar view change enables training of object reaching policies using reinforcement-learning-based sequential object localisation. To facilitate the adaptation of the policy to view changes in the robot workspace, we further present a dynamic filter that learns to bias an input state to remove irrelevant information for an action decision. 
The proposed policy learner can be used as a powerful representation for robotic tasks, and we validate it on static and moving object reaching tasks.
\end{abstract}

%%%%%%%%%%%%%%%%%%%%%%%%%%%%%%%%%%%%%%%%%%%%%%%%%%%%%%%%%%%%%%%%%%%%%%
\section{Introduction}
Data-driven robot learning trains an agent to acquire skills for tasks, such as locomotion~\cite{ratliff2007imitation}, grasping~\cite{peters2008reinforcement}, and manipulation~\cite{ratliff2007imitation,sharma2019third}. 
However, learning the control policies requires a large amount of training data and expert knowledge.
Reinforcement Learning (RL) aims at learning policies by interacting with the environment~\cite{peters2008reinforcement,levine2016end,levine2018learning}. The agent acquires generalised policies while exploring the environment and maximising a reward signal. RL with real-world training requires expensive and potentially unsafe active data collection by robots to gather a large number of experiences from scratch.

To address the limitation of active data acquisition in traditional learning-based approaches, simulation may be employed to learn control policies that are then transferred to the real world. The domain gap between the simulation and real world is addressed by domain adaptation~\cite{rusu2016sim,bousmalis2018using}, 
domain randomisation~\cite{tobin2017domain}, meta-learning~\cite{finn2017model,duan2017one} or learning from depth images~\cite{gualtieri2016high,viereck2017learning,gualtieri2018learning}. 
When the policy trained in simulation is adapted to the real-world robot workspace with sim-to-real techniques, precise camera calibration is needed to ensure the coherency between the geometric information from the simulation and real-world~\cite{gualtieri2018learning,yan2017sim}.
\begin{figure*}[t]
    \centering
    \includegraphics[trim = 1.0cm 5.0cm 1.0cm 6.5cm, clip, width=0.9\textwidth]{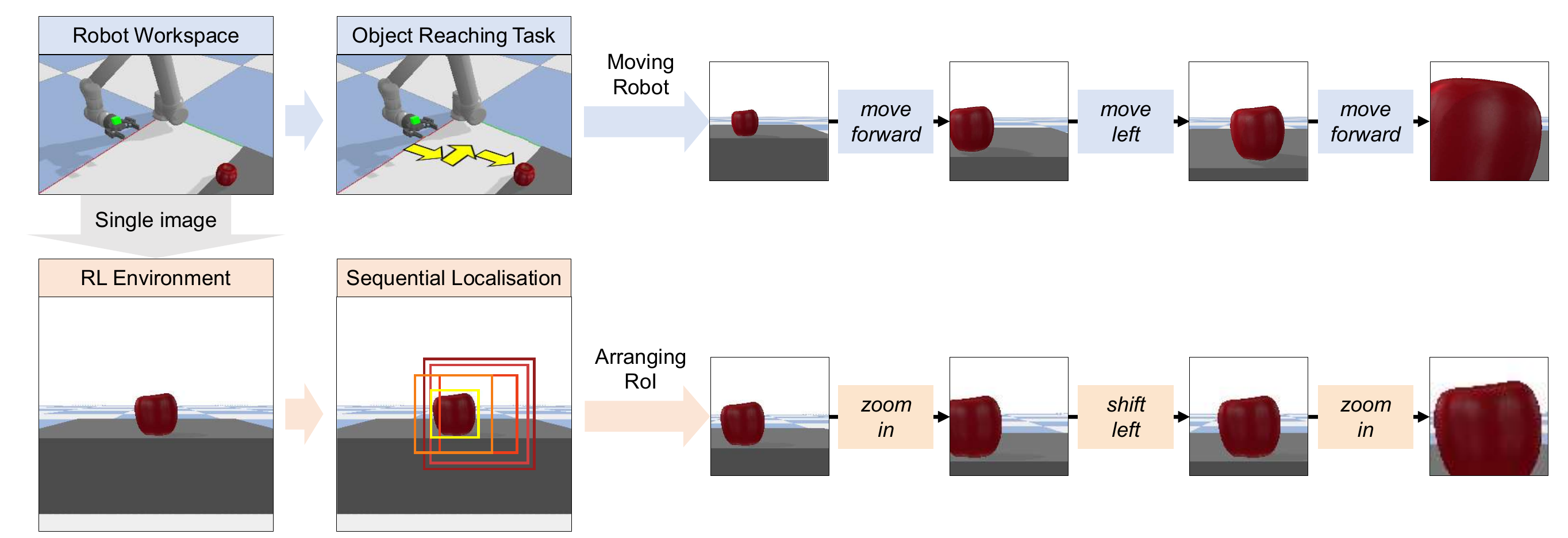}
    \caption{Sequential input states and corresponding generated actions from (top) hand-eye robot manipulation and (bottom) RL-based object localisation. In (top-left) the robot workspace with hand-eye UR5 robot, our method takes (bottom-left) a single image in first-person view from the camera (green cube) on the gripper as an RL environment for sequential object localisation.
    In the RL environment, the input state for the RL agent is produced by cropping and resizing the RoI based on the bounding box. That is, the RL agent sequentially localises the object by arranging the position (shift) and size (zoom) of the RoI with discrete actions. The input states for the sequential localisation task show similar view changes to the hand-eye robot reaching task in the robot workspace. The agent thus can be directly deployed to the robot workspace since the proxy task is performed in the nearly same domain.
    }
    \label{fig:seqandrobot}
\end{figure*}

In this paper, we propose a one-shot hand-eye robot policy learner (OHPL) for the object reaching task. OHPL does not require calibration information nor sim-to-real adaptation. Given the first-person view provided by a camera placed on the end-effector, we observe that the hand-eye robot view changes for the reaching task are similar to locating the object into a region-of-interest (RoI) from the first-person-view image. In OHPL, we first capture a single image from the robot workspace and then use RL-based sequential object localisation as a proxy task to learn the control policy for the object reaching task. OHPL learns the policy to generate sequential actions to change the position and size of the RoI (from the image) until it localises the object. 
To cope with the view changes in the robot workspace, OHPL includes a dynamic filter that generates image-conditioned parameters naturally learned to bias an input state to remove irrelevant information for an action decision. The learned policy can then be deployed for both the static and moving object reaching task in the robot workspace\footnote{Code and video results are available at: \url{http://corsmal.eecs.qmul.ac.uk/OHPL.html}}.

%%%%%%%%%%%%%%%%%%%%%%%%%%%%%%%%
\section{Related Work}

In this section, we review learning-based robot control, the domain gap problem, and proxy tasks to learn representations without explicit supervision.

Control policies can be learned with RL or in a supervised fashion (e.g.,~\textit{imitation learning}, IL).
With IL, agents acquire skills from demonstrations, which include kinesthetic demonstrations~\cite{calinon2007learning}, teleoperation~\cite{pastor2009learning}, and demonstrations from a human~\cite{stadie2017third,sermanet2018time,song2020grasping}. Expert demonstrations can be used directly to train the agent behaviour~\cite{duan2017one,pomerleau1991efficient} or be used to estimate a reward function for RL~\cite{abbeel2004apprenticeship}.
Although IL can achieve good performance using expert demonstrations to guide policy learning~\cite{tai2016survey}, it requires a large amount of training data and policies can be biased to the specific demonstration.

RL supports learning skill policies through trial and error, interactively within an environment by maximising cumulative reward received by the agent. RL can be directly applied in the real world to learn end-to-end policies for motor control~\cite{peters2008reinforcement,levine2016end,levine2018learning}, but a large number of trials are needed to generalise a policy and manual resetting may be needed between episodes, increasing the cost for data collection~\cite{levine2016end,levine2018learning,kalashnikov2018qt}.
Policies are thus often trained first in simulated environments, then transfer to a real-world task~\cite{rusu2016sim,bousmalis2018using,tobin2017domain}. 
If there is no significant domain shift between the simulation and test environments, the learned policy can be directly used in the test environment~\cite{tzeng2020adapting}.
However, transferring skills is challenging due to both visual and dynamics differences between the two domains~\cite{rusu2016sim,bousmalis2018using,tobin2017domain}. 

To reduce the domain gap, a manipulation model can be learned with synthetic objects rendered with real object textures~\cite{saxena2008robotic}. Another approach is to learn a policy from a modality with a relatively small domain gap, e.g. depth images ~\cite{gualtieri2016high,viereck2017learning,gualtieri2018learning}.   
Visual domain adaptation has also been used for robot manipulation, transferring knowledge from simulation to the real world. The methods include transfer learning~\cite{rusu2016sim} and learning policy with generated real-like images~\cite{bousmalis2018using} or simplified images using generative adversarial networks~\cite{james2019sim}.  

Domain randomisation varies the appearance (e.g. textures, lighting and camera position) of sensory inputs during training, thus helping the agent learn relevant features that are invariant across the domain gap~\cite{tobin2017domain}. This technique can be applied to various neural network-based robot manipulation approaches with learning feed-forward networks ~\cite{tobin2017domain,james2017transferring}, imitation learning ~\cite{james2018task,bonardi2020learning}, and reinforcement learning ~\cite{james2019sim,matas2018sim,iqbal2020toward}.
While domain randomisation provides good performance in addressing the domain gap, the trained model still has to address an unseen environment during testing. Domain randomisation in reinforcement learning setting may also lead to stability issues~\cite{matas2018sim}. 
{Since these real-to-simulation approaches do not learn the policy directly from the robot workspace, precise calibration is also a core issue to address robot manipulation~\cite{yan2017sim,gualtieri2018learning}.}

When expert supervision is unavailable or insufficient, a proxy task is an alternative to pre-train representations to be used, after finetuning, for specific tasks. 
For instance, visual representations can be learned from images, without annotations, by solving a proxy task, such as reconstructing input under corruptions such as noise, holes, colour, by performing denoising~\cite{vincent2008extracting}, inpainting~\cite{pathak2016context}, and colourisation~\cite{zhang2016colorful}, respectively. The representations can also be learned by classifying input with pseudo-label such as patch orders~\cite{doersch2015unsupervised}, segmenting object motions~\cite{pathak2017learning}, and clustering images~\cite{caron2018deep}. 
An example proxy task for the RL agent is predicting a future state from the current state and action, which helps the agent to encode the representations that are effective for achieving the goal~\cite{pathak2017curiosity,lesort2018state}.

\section{Learning Object Localisation as a~Proxy~Task}

\begin{figure*}[t]
    \centering
    \subfloat{
     \includegraphics[trim = 4.0cm 7.5cm 4.0cm 7.3cm, clip, width=0.8\textwidth]{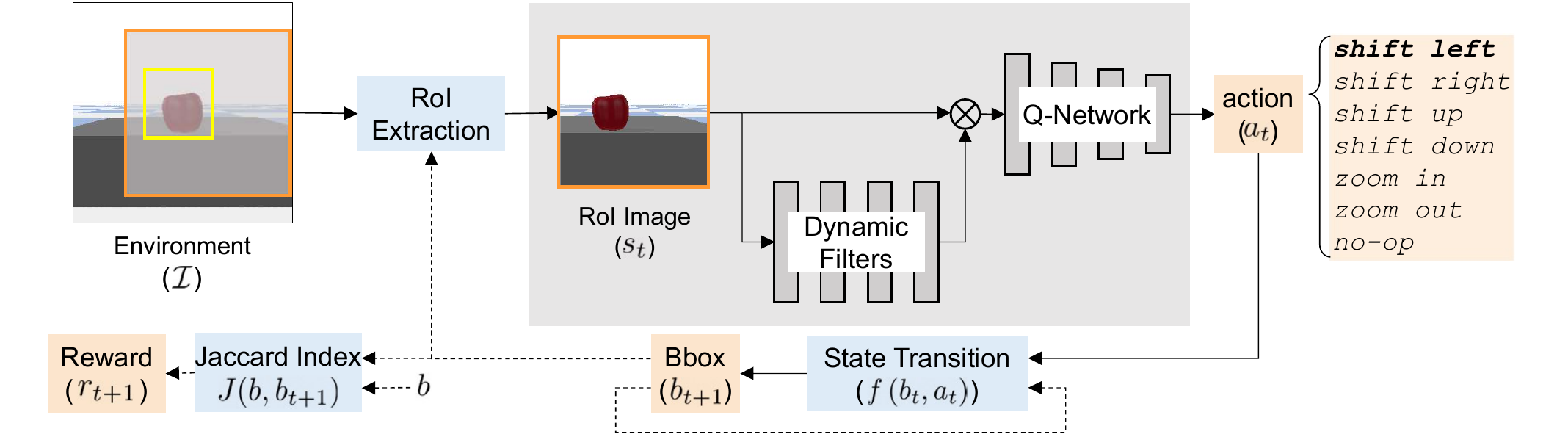}
     }
    \caption{Overview of the one-shot hand-eye policy Learner (OHPL). Given a single image as the RL environment, $\mathcal{I}$, and the bounding box (Bbox) at time step $t$, $b_t$, the agent takes an extracted and resized RoI as an input state, $s_t$, passes it through the dynamic filter and then performs element-wise multiplication ($\otimes$) with $s_t$. The combined input is fed into the Q-Network to determine the action, $a_t$. Based on $a_t$, the size or position of the next RoI, $b_{t+1}$, is changed with the state transition, $f\left(b_t,a_t\right)$ to locate the target object. For instance, the agent generates $a_t$ as \textit{shift left} to move $b_t$ (outlined orange) towards the groundtruth bounding box (outlined yellow), $b$, in the next step.
    The process is repeated until the Jaccard index between $b$ and the current bounding box becomes higher than 0.8 or the agent moves the maximum actions in an episode. The dotted line denotes the process for $t+1$ step. The process in the grey box can be directly deployed to the robot workspace where the hand-eye robot performs sequential discrete actions.}
    \label{fig:overview}
\end{figure*}
\subsection{Problem definition}

We cast the reaching task in the robot workspace as a sequential object localisation task in a first-person-view image~\cite{yun2017action,jie2016tree}. The view is changed by a set of discrete {\em robot actions} (Fig.~\ref{fig:seqandrobot}, top). These view changes are similar to sequential object localisation with discrete {\em image-processing actions} (Fig.~\ref{fig:seqandrobot}, bottom).
We hypothesise that the model trained with such a task can, if appropriately formulated, naturally acquire skills for the agent to perform the reaching task in the robot workspace.

We formulate the sequential object localisation task by locating the RoI as a Markov Decision Process (MDP). We take an initial observation from the robot, a single image $\mathcal{I}$, to form an environment for RL. 
In the MDP, the RL agent learns to change the position and size of the bounding box over some discrete time steps until the RoI image captures the object in $\mathcal{I}$.
At each time step $t$, the agent receives a state, $s_t$, the RoI image resized to the network input size, and selects an action, $a_t$, among $N$ possible actions, according to the policy that maps $s_t$ to $a_t$. 
After executing the action, a state transition function generates the next state, $s_{t+1}$ and the agent receives a reward $r_t$.

\subsection{State and action}
The overview of our approach is shown in Fig.~\ref{fig:overview}.
At time step $t$, a square bounding box is represented by a three-dimensional vector $b_t=\left[x_t, y_t, w_t  \right]$ where $\left[x_t, y_t\right]$ and $w_t$ denote the top-left pixel and the window size of the bounding box, respectively. 
The RoI image, based on the bounding box, is extracted from the RL environment, $\mathcal{I}\in {\mathbb{R}^{360 \times 360 \times 3}}$, and then resized to the network input size, which becomes the input state $s_t\in {\mathbb{R}^{84 \times 84 \times 3}}$ to the agent.
The agent consists of convolutional neural networks that reduce dimension of the high-dimensional input state~\cite{pathak2017curiosity,mnih2013playing}.
We set the initial state, $s_1$, to an RoI image that includes the object but with random position and size of the bounding box. This random spawning strategy enables the agent to learn the generalised localisation policy in the single image environment $\mathcal{I}$.

We consider seven discrete robot actions ($N=7$): \textit{move left}, \textit{move right}, \textit{move up}, \textit{move down}, \textit{move forward}, \textit{move backward}, and \textit{stop}. 
These actions and corresponding hand-eye view changes can be imitated in the proxy task that changes the position and size of the RoI in $\mathcal{I}$ with the following discrete actions: \textit{shift left}, \textit{shift right}, \textit{shift up}, \textit{shift down}, \textit{zoom in}, \textit{zoom out}, and \textit{no-operation (no-op)}.
For instance, moving forward and left of the robot corresponds to \textit{zoom in} and \textit{shift left} actions by changing the window size, $w_t$, of the RoI and its position.

After the action $a_{t}$ is decided from $s_t$, the next state $s_{t+1}$ is obtained by the state transition function $f\left(b_t,a_t\right)$ that arranges the position and size of the current bounding box, $b_t$. By arranging the position $\left(x_t, y_t\right)$ with the displacement $\left(\Delta{x_t}, \Delta{y_t}\right)$, and $w_t$ with the RoI size change $ \Delta{w_t}$, we obtain the next bounding box, $b_{t+1}$ as follows:
\begin{equation}
    b_{t+1}=\left[x_t+\Delta{x_t}, y_t+\Delta{y_t}, w_t+\Delta{w_t}  \right].
\end{equation}
The amount of displacement depends on the size of the RoI, which corresponds to the view changes in the robot workspace.
For example, the displacement between the current and next state is small when the object is close, while the displacement is large when the object is far. 
Therefore, we adapt the amount of displacement based on the RoI size as follows:
\begin{equation}\label{eq:transition}
\Delta{x_t} = {\sigma}_t {w_t}, \Delta{y_t} = {\sigma}_t {w_t},
\end{equation}
where ${\sigma}_t$ controls the ratio between the amount of displacement and the bounding box size. 
Considering the PID control error that is likely to occur in real robot control we randomly generate ${\sigma}_t \in [0.05,0.15]$ in every step. 
The agent thus can learn policies with more generalised cases to reach the goal position. 
The next state $s_{t+1}$ can be finally obtained by extracting the region based on the computed $b_{t+1}$ and resizing. For exception handling, we set $s_{t+1} = s_{t}$ when the location of $b_{t+1}$ is outside of $\mathcal{I}$ or $w_t$ is smaller or larger than 20 and 360, respectively.

\begin{figure*}[t!]
    \centering
    
    \subfloat{
    \includegraphics[width=0.13\linewidth]{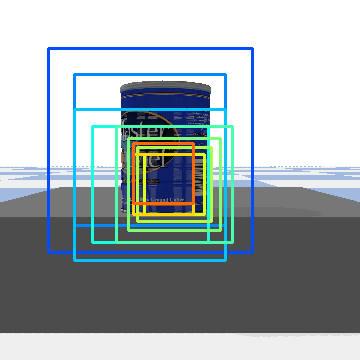}} \hspace{9pt}
    \subfloat{
     \includegraphics[width=0.08\linewidth]{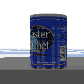}
     \includegraphics[width=0.08\linewidth]{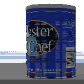}
     \includegraphics[width=0.08\linewidth]{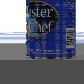}
     \includegraphics[width=0.08\linewidth]{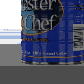}
     \includegraphics[width=0.08\linewidth]{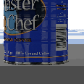}
     \includegraphics[width=0.08\linewidth]{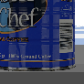}
     \includegraphics[width=0.08\linewidth]{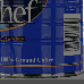}
     \includegraphics[width=0.08\linewidth]{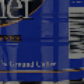}}\\
    \caption{Sequential object localisation examples in (a) the RL environment $\mathcal{I}$, where the bounding box is colour-coded from dark blue to yellow to indicate the temporal evolution of the actions, and (b) corresponding sequence of input states for the OHPL agent. }
    \label{fig:localisation} \vspace{-9pt}
\end{figure*}

\begin{figure*}[t!]
    \centering
    \subfloat[Gaussian blur]{\includegraphics[trim = 0.5cm 1.5cm 1.0cm 3cm, clip, width=0.33\linewidth]{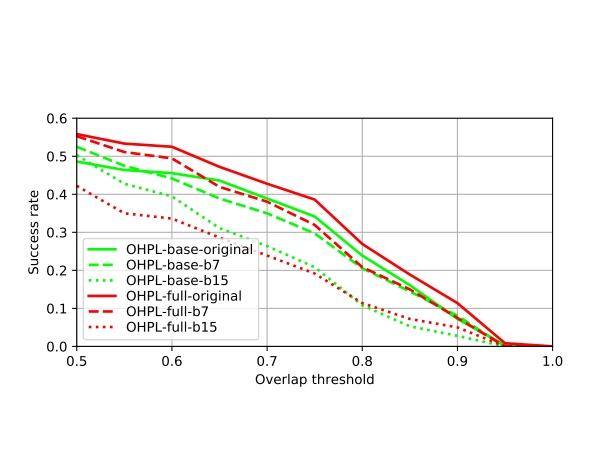}}
    \subfloat[Gaussian noise]{\includegraphics[trim = 0.5cm 1.5cm 1.0cm 3cm, clip, width=0.33\linewidth]{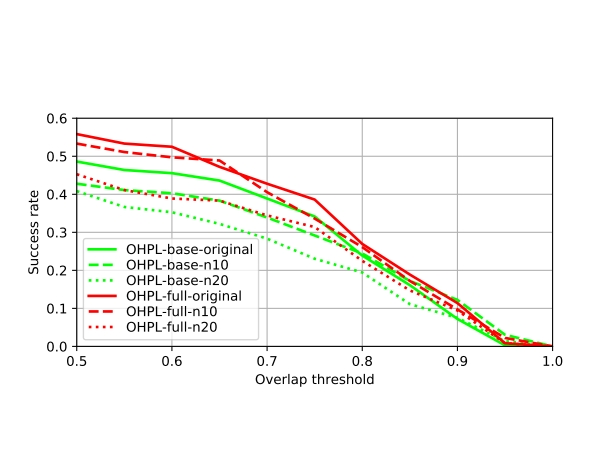}}
    \subfloat[Illumination]{\includegraphics[trim = 0.5cm 1.5cm 1.0cm 3cm, clip, width=0.33\linewidth]{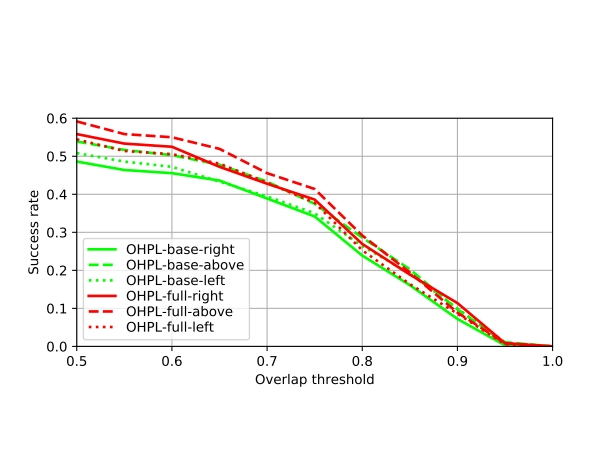}}
    \caption{Sequential object localisation results when varying the overlap thresholds on the test data. To evaluate the model robustness, we tested the sequential object localisation by corrupting the original image by applying (a) Gaussian blur, (b) Gaussian noise (c) and different location of the light source. }
    \label{fig:oneshotratio}
\end{figure*}

\subsection{Reward function}

We devise a reward function that encourages an RL agent to localise the object by producing dense rewards. The reward function integrates intrinsic
motivations to encourage the agent to naturally move towards the goal to achieve higher Jaccard index, alleviating the problem of sparse extrinsic rewards. 
For each time step $t$, the reward, $r_t$, is defined as follows: 
\begin{equation}\label{eq:extrinsic}
r_t=
\begin{cases}
1 & J\left(b, b_t\right)>0.8\\
\alpha\left(J\left(b, b_t\right)-1\right) &{0.8\geq 
J\left(b,b_t\right)>0}\\
-$1$ & J\left(b, b_t\right)=0,
\end{cases}   
\end{equation}
where $J\left(b,b_t\right)$ measures the overlap ratio between the groundtruth bounding box, $b$, and the bounding box at $t$, $b_t$. We consider that the agent has reached the goal when the overlap ratio between $b$ and $b_t$ is higher than 0.8, where the agent receives a positive reward of $+1$. 
Instead of giving a reward of $-1$ to all the states which fail to achieve the goal, we introduce a small negative reward that encourages the agent to move in order to localise the object with higher Jaccard index score. When $b_t$ partially overlaps with $b$, the small negative reward can be obtained by $\alpha\left(J\left(b,b_t\right)-1\right)$, with $\alpha = 0.5$. The higher the Jaccard index, the smaller the negative reward.
Lastly, a negative reward of $-1$ is given when $b_t$ does not overlap with $b$. This reward setting encourages the RoI to not move away from the object during training. The episode is terminated when the agent reaches the goal or has reached the maximum number of steps $t=T_{max}$.

\subsection{Learning policies with Deep Q-Network}
We address the learning of the proxy task using the DQN framework~\cite{mnih2013playing}. We employ a state encoding network to approximate the underlying action-value function:
\begin{equation}\label{eq:dqn}
Q(s_t,a_t)=
r_t + \gamma \max_{a_{t+1}}Q\left(s_{t+1},a_{t+1}\right),
\end{equation}
where $\gamma$ is a discount factor that indicates the value of future rewards in the current state. The Q-value $Q\left(s_{t+1},a_{t+1}\right)$ presents how good it is to select $a_{t+1}$ at $s_{t+1}$ to receive a high reward. $Q\left(s_{t+1},a_{t+1}\right)=0$ when $s_{t+1}$ is the terminal state.
Deep reinforcement learning methods commonly require a large discount factor as the agent may explore thousands of steps before reaching the terminal target. 
The goal in our environment is around 7-15 steps far away, and we set a low discount factor, $\gamma=0.85$ to ensure that the agent follows the intrinsic motivations while optimising for long term success.

Note that discrepancies exist between inputs generated from the 2D single image, $\mathcal{I}$, and the 3D robot workspace, due to sensor noise, blur, illumination, and object shape changes. 
We propose to use a dynamic filter~\cite{jia2016dynamic} as soft-attention to provide effective information as an input for localisation.
The network takes $s_{t}$ as an input and generates the feature map which has the same size as $s_{t}$. We then fuse the output feature map and $s_{t}$ by multiplying them in an element-wise manner, where the output becomes an input for the Q-Network. Noting that the feature map generated from the dynamic filter is conditioned on the input, the dynamic filter naturally learns to bias the input during the training without any loss function. This process can be considered as estimating the soft-attention based on the input image.

Fig.~\ref{fig:localisation} shows an example of sequential object localisation. Initially, a bounding box is set with a large RoI, and then the agent arranges the RoI position and size to locate the object. 

\begin{figure*}[!t]
    \centering
    \includegraphics[width=0.10\linewidth]{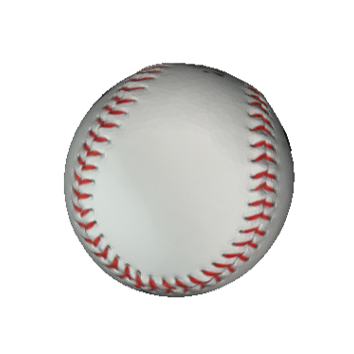}
    \includegraphics[width=0.10\linewidth]{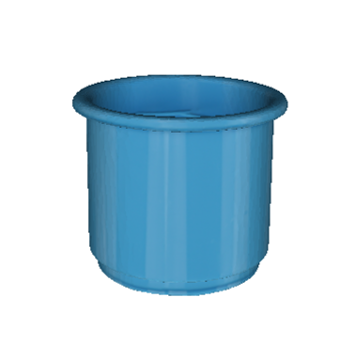}
    \includegraphics[width=0.10\linewidth]{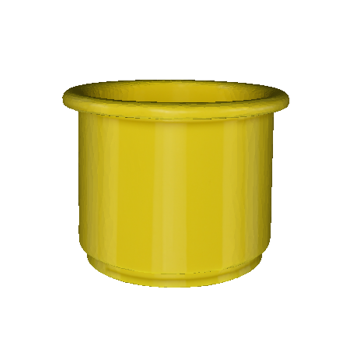}
    \includegraphics[width=0.10\linewidth]{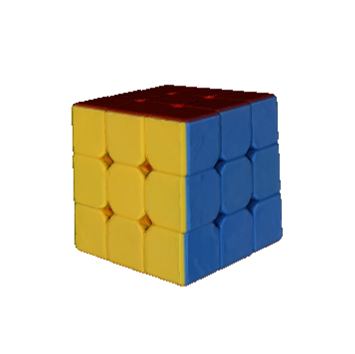}
    \includegraphics[width=0.10\linewidth]{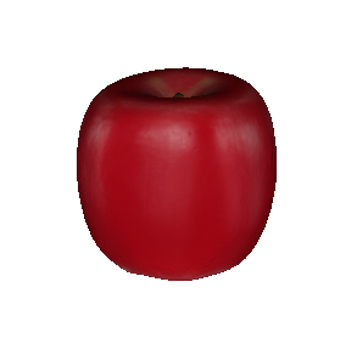}
    \includegraphics[width=0.10\linewidth]{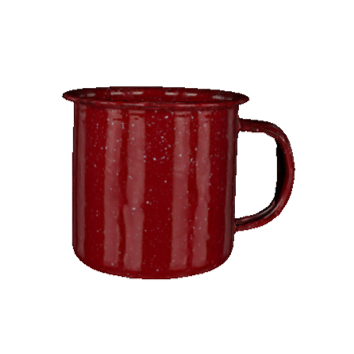}
    \includegraphics[width=0.10\linewidth]{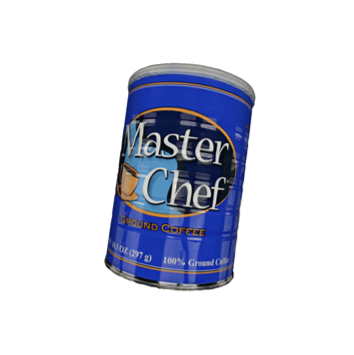}
    \includegraphics[width=0.10\linewidth]{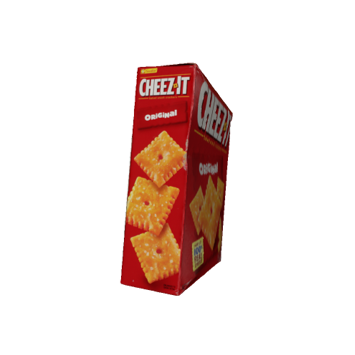}
    \\
    
    \includegraphics[width=0.10\linewidth]{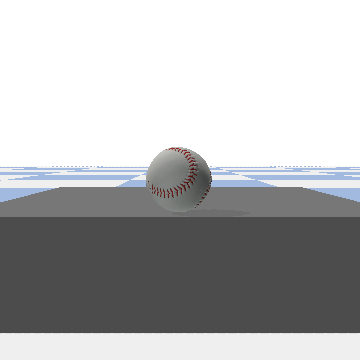}
    \includegraphics[width=0.10\linewidth]{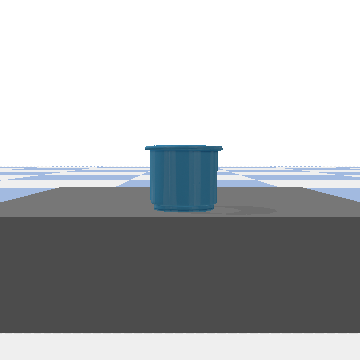}
    \includegraphics[width=0.10\linewidth]{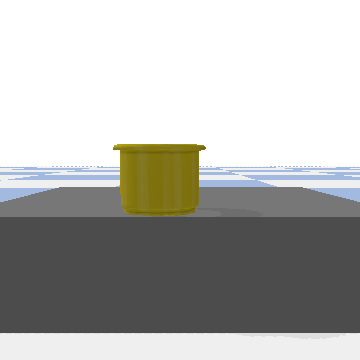}
    \includegraphics[width=0.10\linewidth]{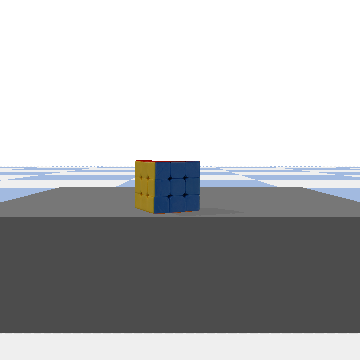}
    \includegraphics[width=0.10\linewidth]{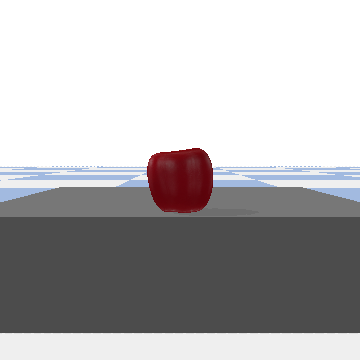}
    \includegraphics[width=0.10\linewidth]{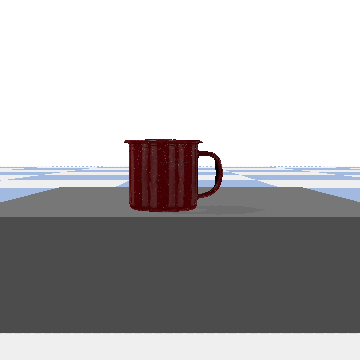}
    \includegraphics[width=0.10\linewidth]{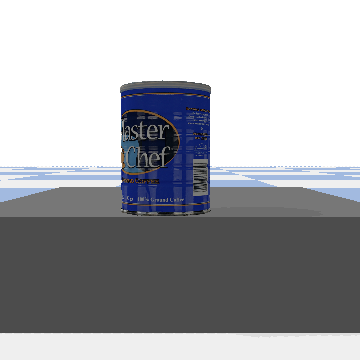}
    \includegraphics[width=0.10\linewidth]{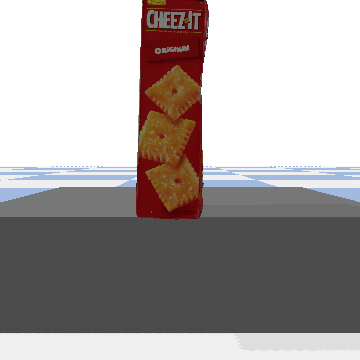}
    \\
    \caption{(Top) Test objects from YCB Benchmark~\cite{Calli2015YCB} and (bottom) their first-person view as an RL environment, $\mathcal{I}$, for training the OHPL agent. }
    \label{fig:ycb}
\end{figure*}

%%%%%%%%%%%%%%%%%%%%%%%%%
\section{Validation}

\subsection{Models under analysis}

We evaluate the OHPL agent on the reaching task when deployed in the robot simulation workspace, PyBullet~\cite{coumans2019pybullet}, but first we analyse the proposed proxy task, RL-based sequential object localisation in the single image environment using two models, namely OHPL-base (vanilla DQN model~\cite{mnih2013playing}), and OHPL-full (DQN model with dynamic filter). A single image for each object illuminated from the left is used as the RL environment (see Fig.\ref{fig:ycb}). We use 8 synthetic objects, i.e. \textit{baseball}, \textit{cup-H}, \textit{cup-J}, \textit{rubiks cube}, \textit{apple}, \textit{mug}, \textit{can}, \textit{cracker box}, that are adopted from the Yale-Carnegie Mellon University-Berkeley (YCB) benchmark~\cite{Calli2015YCB}, including objects with different shapes, textures, and colours.  As each image is employed as a separate RL environment, we generate 8 models each on OHPL-base and OHPL-full, respectively.
In each episode, we randomly set a position and size for the initial RoI, while including the object in the RoI.
We set $T_{max}=50$ for a single episode, where the localisation can be achieved in 7-15 steps.
We set the position of the ground-truth bounding box as shown in the rightmost images in Fig.~\ref{fig:localisation}. 
We compare OHPL-base and OHPL-full on the test environment with nine object locations, \textit{middle (M)}, \textit{mid-left (ML)}, \textit{mid-right (MR)}, \textit{top (T)}, \textit{top-left (TL)}, \textit{top-right (TR)}, \textit{bottom (B)}, \textit{bottom-left (BL)}, and \textit{bottom-right (BR)}, and three different illuminations by setting the light source location to left, right, and above. We perform 10 trials for each object location and illumination.

We further test the models by corrupting the test images with Gaussian noise, Gaussian blur, and varying illumination condition, which emulates camera sensor noise, blur, illumination changes in the robot workspace.
The Gaussian blur is applied with two kernel sizes, 7 and 15, and the Gaussian noise with two variances, 10 and 20. Noting that the light in the training environment is located to the left-hand side, we also change the light directions to right and above. 

We then validate the OHPL agent, trained from the proxy task, in the robot simulation workspace for the object reaching task. We use a 6 DoF UR5 robotic arm~\cite{ur5} with the Robotiq 2F-85 gripper~\cite{gripper}. The arm is controlled by specifying a pose goal in the Cartesian space. Actions issued to the arm have a fixed displacement value of 0.02m. The object is placed on a platform in front of the gripper at about 30 steps away. $360\times360\times3$ images are captured at each step. We consider the task completed when the gripper reaches a predefined area in front of the object.

We compare the OHPL agent, {\em without robot training}, and a model {\em with} robot training. For the latter, we {\em train an agent from scratch} (Scratch) following the DQN framework~\cite{mnih2013playing}. The Scratch agent first explores randomly for 1000 steps and then follows the $\epsilon$-greedy strategy. We pre-defined the explorable region to prevent the agent from exploring states irrelevant to the task and to prevent the robot from colliding with the environment e.g., a grey box to place the target object. We set the maximum steps in an episode as $T_{max} = 50$ and give a positive extrinsic reward when successful reaching is performed. In addition to the extrinsic reward, at each step, an intrinsic reward is given based on the distance between the gripper and the object to encourage the robot to reach the object and converge faster.

Finally, we validate the OHPL agent by deploying the trained agents to the moving object reaching task where the environment is unseen to both the OHPL agent and the agent with robot training (Scratch). We initialise the object floating without the grey box in the environment. During each step, we randomise the velocity of the object in the x, y and z directions. The same success criteria from the static object reaching task is used. Experiments are performed on all 8 objects in 3 lighting conditions and 9 starting positions. 3 trials are performed for each combination so a total of 648 experiments are performed for each model.

\subsection{Network setting}
% When training an OHPL agent for sequential object localisation, the input state is cropped from the image environment $\mathcal{I}$ based on the RoI and 
In this section, we introduce the network configuration for OHPL agent. For sequential object localisation, an input state is cropped from the RL environment $\mathcal{I}$ based on the RoI. The hand-eye view is used as the input state for object reaching task in the robot workspace. For both tasks, the input state is then resized to $84\times84\times3$. We adopt the context aggregation network~\cite{yu2015multi} architecture with the rectified linear unit (ReLU) as nonlinear activation for the dynamic filter using 4 convolution layers with 3 intermediate feature maps. We set the filter kernels size to $3\times3$.
The last convolution layer applies a $1\times1$ convolution followed by a sigmoid to predict the final output, which is then combined with the RoI. The dilation factors of the convolution layers are 1, 2, 4, and 1, respectively. 
As Q-Network we use the vanilla DQN~\cite{mnih2013playing} with minor modification to the final fully connected layer with a single output for each valid action. The experience replay has a $5k$ memory volume. Note that the dynamic filter makes 0.01$\%$ parameter increase to the total number of parameters of the network.
The network is optimised by RMSprop~\cite{tieleman2012lecture}, where the learning rate is set to $0.001$ and decays exponentially. We adopt the $\epsilon$-greedy strategy during training and exponentially anneal $\epsilon$ from 0.95 to 0.05~\cite{mnih2013playing}.

%%%%%%%%%%%%%%%%%%%%%%%%%%%%%%%%%%%%%%%%%%%%%%%%%%%%%%%%%%%%%%%%%%
%https://personal.sron.nl/~pault/data/colourschemes.pdf
\definecolor{c1}{RGB}{0,68,136}  % Blue
\definecolor{c2}{RGB}{51,187,238} % CYAN
\definecolor{c3}{RGB}{17,119,51}  % GREEN
\definecolor{c4}{RGB}{238,119,51} % ORANGE
\definecolor{c5}{RGB}{204,51,17} % RED
\definecolor{c6}{RGB}{238,51,119} % Magenta
\definecolor{c7}{RGB}{187,187,187} % Gray
\definecolor{c8}{RGB}{0 0 0} % Black
\definecolor{colorR}{RGB}{204, 38, 38}
\definecolor{colorG}{RGB}{69, 160, 37}
\definecolor{colorB}{RGB}{68, 114, 196}

%%%%%%%%%%%%%%%%%%%%%%%%%%%%%%%%%%%%%%%%%%%%%%%5

\begin{table}[t!]
\caption{Success ratio, with standard error in brackets, for the static object reaching task in the robot workspace with and without training in the robot workspace (Robot.), and with (Arch. full) and without (Arch. base) the dynamic filter  }
\label{tab:static}
\vspace{-9pt} \vspace{-9pt}
\begin{center}
\footnotesize 
\setlength\tabcolsep{1 pt}
\renewcommand\theadset{\def\arraystretch{.5}}%
\begin{tabular}{lcc  ccc ccc ccc}
\Xhline{1.0pt}
\multicolumn{3}{c}{\textbf{Model}}  & \multicolumn{9}{c}{\textbf{Initial position}}     \\
\cmidrule(lr){1-3}
\cmidrule(lr){4-12}
{\textit{Method}} & \textit{Robot.} & {\textit{Arch.}} &   \textit{M} & \textit{ML} & \textit{MR} & \textit{T} & \textit{TL} & \textit{TR} & \textit{B} & \textit{BL} & \textit{BR}  \\
\Xhline{1.0pt}       
\multirow{4}{*}{\thead{Scratch}}       &  \checkmark    &  base   &{\thead{100.0 \\ (0.0)}} &{\thead{100.0 \\ (0.0)}} &{\thead{87.5 \\ (6.8)}} &{\thead{100.0 \\ (0.0)}} &{\thead{91.7 \\ (5.6)}} &{\thead{100.0 \\ (0.0)}} &{\thead{100.0 \\ (0.0)}} &{\thead{100.0 \\ (0.0)}} &{\thead{95.8 \\ (4.1)}}    \\
&  \checkmark    &  full   &{\thead{100.0 \\ (0.0)}} &{\thead{100.0 \\ (0.0)}} &{\thead{100.0 \\ (0.0)}} &{\thead{100.0 \\ (0.0)}} &{\thead{100.0 \\ (0.0)}} &{\thead{87.5 \\ (6.8)}} &{\thead{87.5 \\ (6.8)}} &{\thead{100.0 \\ (0.00)}} &{\thead{87.5 \\ (6.8)}} \\
\midrule
\multirow{4}{*}{{OHPL}}    &   & base &  {\thead{95.8 \\ (4.1)}} &{\thead{95.8 \\ (4.1)}} &{\thead{75.0 \\ (8.8)}} &{\thead{91.7 \\ (5.6)}} &{\thead{95.8 \\ (4.1)}} &{\thead{70.8 \\ (9.3)}} &{\thead{20.8 \\ (8.3)}} &{\thead{8.3 \\ (5.6)}} &{\thead{4.2 \\ (4.1)}}    \\

&  & full   &   {\thead{95.8 \\ (4.1)}} &{\thead{79.2 \\ (8.3)}} &{\thead{79.2 \\ (8.3)}} &{\thead{79.2 \\ (8.3)}} &{\thead{87.5 \\ (6.8)}} &{\thead{83.3 \\ (7.6)}} &{\thead{37.5 \\ (9.9)}} &{\thead{54.2 \\ (10.2)}} &{\thead{4.2 \\ (4.1)}}    \\

\Xhline{1.0pt}
\end{tabular}
\end{center}
 
\vspace{-9pt}
\end{table}
%
%%%%%%%%%%%%%%%%%%%%%%%%%%%%%%%%%%%%%%%%%%%%%%%%%
% Reaching plot data
\pgfplotstableread{055_baseball_reaching_scratch.txt}\reachingbaseballscratch
\pgfplotstableread{065-h_cups_reaching_scratch.txt}\reachingcupHscratch
\pgfplotstableread{065-j_cups_reaching_scratch.txt}\reachingcupJscratch
\pgfplotstableread{077_rubiks_cube_reaching_scratch.txt}\reachingrubiksscratch
\pgfplotstableread{013_apple_reaching_scratch.txt}\reachingapplescratch
\pgfplotstableread{025_mug_reaching_scratch.txt}\reachingmugscratch
\pgfplotstableread{002_master_chef_can_reaching_scratch.txt}\reachingcanscratch
\pgfplotstableread{003_cracker_box_reaching_scratch.txt}\reachingcrackerscratch

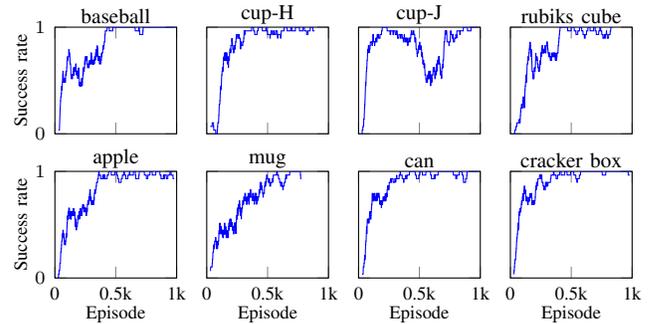
\begin{figure}[t]
    \centering
    \centering
    
    \begin{tikzpicture}
    \begin{groupplot}[group style={group size=4 by 2, horizontal sep=0.4cm, vertical sep=0.5cm}]
    
    \nextgroupplot[%
    width=3.2cm,
    height=3cm,
    xmin=0, xmax=1000,
    xtick={0, 500, 1000},
    xticklabels={},
    ymin=0.0, ymax=1.0,
    ytick={0, 1.0},
    ylabel = {Success rate},
    ylabel style = {at={(-0.15,0.5)},font=\scriptsize},
    tick label style={font=\scriptsize},
    title={baseball},
    title style={at={(0.5,0.8)},font=\footnotesize}]
    \addplot+[solid, color=blue, mark=none] table[x=x_val, y=y_val]{\reachingbaseballscratch};
    
    \nextgroupplot[%
    width=3.2cm,
    height=3cm,
    xmin=0, xmax=1000,
    xtick={0, 500, 1000},
    xticklabels={},
    ymin=0.0, ymax=1.0,
    ytick={0, 1.0},
    yticklabels={},
    tick label style={font=\scriptsize},
    title={cup-H},
    title style={at={(0.5,0.8)},font=\footnotesize}]
    \addplot+[solid, color=blue, mark=none] table[x=x_val, y=y_val]{\reachingcupHscratch};
    
    \nextgroupplot[%
    width=3.2cm,
    height=3cm,
    xmin=0, xmax=1000,
    xtick={0, 500, 1000},
    xticklabels={},
    ymin=0.0, ymax=1.0,
    ytick={0, 1.0},
    yticklabels={},
    tick label style={font=\scriptsize},
    title={cup-J},
    title style={at={(0.5,0.8)},font=\footnotesize}]
    \addplot+[solid, color=blue, mark=none] table[x=x_val, y=y_val]{\reachingcupJscratch};
    
    \nextgroupplot[%
    width=3.2cm,
    height=3cm,
    xmin=0, xmax=1000,
    xtick={0, 500, 1000},
    xticklabels={},
    ymin=0.0, ymax=1.0,
    ytick={0, 1.0},
    yticklabels={},
    tick label style={font=\scriptsize},
    title={rubiks cube},
    title style={at={(0.5,0.8)},font=\footnotesize}]
    \addplot+[solid, color=blue, mark=none] table[x=x_val, y=y_val]{\reachingrubiksscratch};
    
    \nextgroupplot[%
    width=3.2cm,
    height=3cm,
    xmin=0, xmax=1000,
    xtick={0, 500, 1000},
    xticklabels={0, 0.5k, 1k},
    xlabel = {Episode},
    xlabel style = {at={(0.5,-0.17)},font=\scriptsize},
    ymin=0.0, ymax=1.0,
    ytick={0, 1.0},
    ylabel = {Success rate},
    ylabel style = {at={(-0.15,0.5)},font=\scriptsize},
    tick label style={font=\scriptsize},
    title={apple},
    title style={at={(0.5,0.8)},font=\footnotesize}]
    \addplot+[solid, color=blue, mark=none] table[x=x_val, y=y_val]{\reachingapplescratch};
    
    \nextgroupplot[%
    width=3.2cm,
    height=3cm,
    xmin=0, xmax=1000,
    xtick={0, 500, 1000},
    xticklabels={0, 0.5k, 1k},
    xlabel = {Episode},
    xlabel style = {at={(0.5,-0.17)},font=\scriptsize},
    ymin=0.0, ymax=1.0,
    ytick={0, 1.0},
    yticklabels={},
    tick label style={font=\scriptsize},
    title={mug},
    title style={at={(0.5,0.8)},font=\footnotesize}]
    \addplot+[solid, color=blue, mark=none] table[x=x_val, y=y_val]{\reachingmugscratch};
    
    \nextgroupplot[%
    width=3.2cm,
    height=3cm,
    xmin=0, xmax=1000,
    xtick={0, 500, 1000},
    xticklabels={0, 0.5k, 1k},
    xlabel = {Episode},
    xlabel style = {at={(0.5,-0.17)},font=\scriptsize},
    ymin=0.0, ymax=1.0,
    ytick={0, 1.0},
    yticklabels={},
    tick label style={font=\scriptsize},
    title={can},
    title style={at={(0.5,0.8)},font=\footnotesize}]
    \addplot+[solid, color=blue, mark=none] table[x=x_val, y=y_val]{\reachingcanscratch};
    
    \nextgroupplot[%
    width=3.2cm,
    height=3cm,
    xmin=0, xmax=1000,
    xtick={0, 500, 1000},
    xticklabels={0, 0.5k, 1k},
    xlabel = {Episode},
    xlabel style = {at={(0.5,-0.17)},font=\scriptsize},
    ymin=0.0, ymax=1.0,
    ytick={0, 1.0},
    yticklabels={},
    tick label style={font=\scriptsize},
    title={cracker box},
    title style={at={(0.5,0.8)},font=\footnotesize}]
    \addplot+[solid, color=blue, mark=none] table[x=x_val, y=y_val]{\reachingcrackerscratch};
    
    % \caption{Object reaching}
    \end{groupplot}
    
    \end{tikzpicture}
    
    \caption{Training the agent from scratch (Scratch-base) in the robot workspace. We report the success rates (vertical axis) with a running average over past 30 episodes for each episode (horizontal axis) on 8 YCB objects. }
    \label{fig:gazeboratio}
\end{figure}
%%%%%%%%%%%%%%%%%%%%%
%
\begin{figure}[!]
    \centering
    \includegraphics[width=0.11\textwidth]{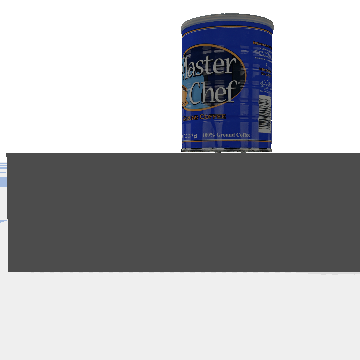}
     \includegraphics[width=0.11\textwidth]{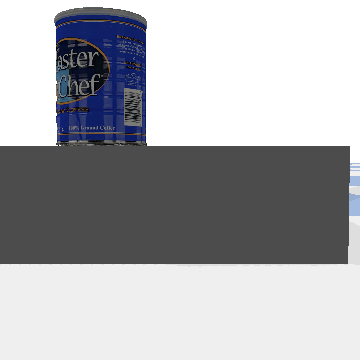}
     \includegraphics[width=0.11\textwidth]{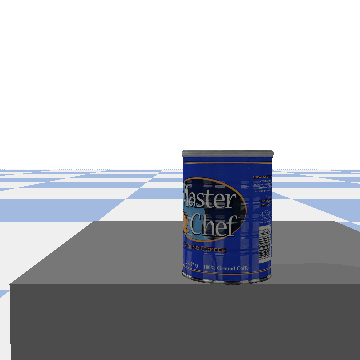}
     \includegraphics[width=0.11\textwidth]{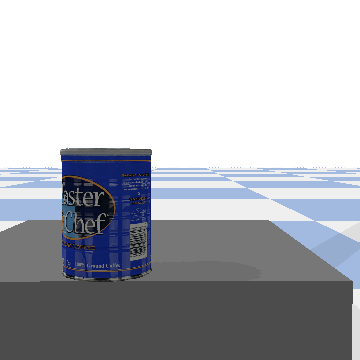}
     \\
     \includegraphics[width=0.11\textwidth]{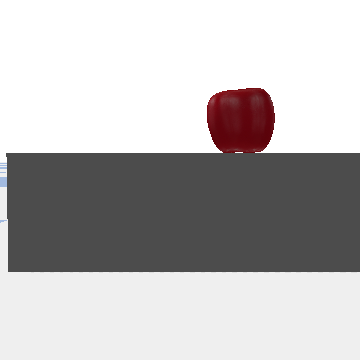}
     \includegraphics[width=0.11\textwidth]{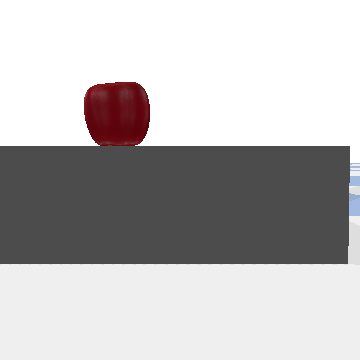}
     \includegraphics[width=0.11\textwidth]{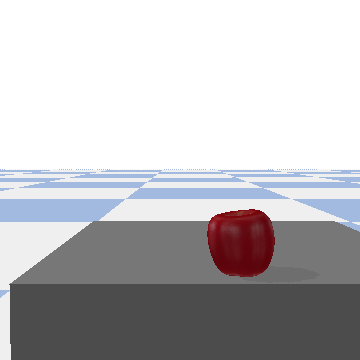}
     \includegraphics[width=0.11\textwidth]{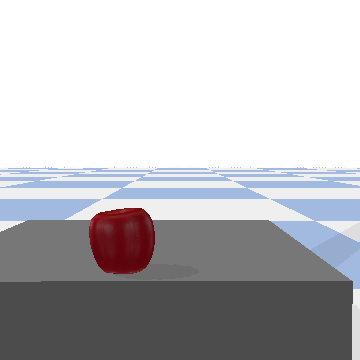}
     \\
    \caption{Examples of hand-eye view from different initial positions, (from left to right) bottom-left (BL), bottom-right (BR), top-left (TL), and top-right (TR). The view from BL and BR is extremely different from the RL environment $\mathcal{I}$ (Fig.~\ref{fig:ycb}), which results in low success of OHP agents. }
    \label{fig:lim}
    \vspace{-9pt}
\end{figure}
%%%%%%

\subsection{Discussion}

Fig.~\ref{fig:oneshotratio} shows the results for {\em sequential object localisation} with varying the image conditions. OHPL-full generally outperforms OHPL-base in localising the object, and the model shows consistent performance in noisy and blurry images. But the performance drops when Gaussian blurring is applied with a $15\times15$ kernel. Also, the performance decreases when the light source is located to right, demonstrating that the performance can be affected when significant light change happens.

Table~\ref{tab:static} shows the evaluation of the {\em static object reaching task} in the robot workspace. Note that Scratch is the upper bound of OHPL as the agent of Scratch models are trained directly in the robot workspace with robot actions. Scratch models are hence overfitted to the environment as the models have experiences with the real view changes and rewards in the robot workspace. However, to learn the control policies for the object-reaching task, each Scratch model requires more than 500 episodes of training in the robot workspace as shown in Fig.~\ref{fig:gazeboratio}. The OHPL agents, without directly learning from a robot workspace and without robot actions, show good results when the initial position is top, \textit{T}, \textit{TL} and \textit{TR}, or middle, \textit{M}, \textit{ML} and \textit{MR}, while showing low success rates in bottom cases, \textit{B}, \textit{BL}, and \textit{BR}.
As shown in Fig.~\ref{fig:lim}, the hand-eye view from the bottom position is extremely different from the middle view which is used as the RL environment $\mathcal{I}$ (see Fig.~\ref{fig:ycb}). OHPL-full, with the additional dynamic filter, records better success rates than OHPL-base on average, by achieving better performance in bottom cases.

For a fair comparison, we further evaluate the agents when they are deployed to the {\em moving object reaching task}. Note that this task can be considered as an unseen scenario to both OHPL and Scratch models. Table~\ref{tab:moving} shows the success ratio of the models. Compared to the success ratio of the static object reaching task, Scratch models show significant performance decrease when they are deployed to the moving object reaching task. While OHPL agents trained their control policies within a single view captured from the static object reaching environment, they generally show good performance and are competitive to the Scratch agents.

%%%%%%%%%%%%%%%%%%%%%%%%%%%%%%%%%%%%%
\section{Conclusion}
We presented a new approach to learn robot control policies with a proxy task: RL-based object localisation with a single image captured from the robot workspace. 
The key idea is that view changes of the RoI in a 2D image are similar to the changes of the hand-eye camera view in the robot workspace. 
We showed that the policy learned from OHPL can be deployed to the robot workspace {\em directly} as the control policy for the reaching task, without finetuning. 
We also presented a dynamic filter that works as soft attention of an input image. With only 0.01\% parameter increase, the network has better generalisation when the agent is deployed to perform static and moving object reaching task. 

Future work includes validating the proposed framework with a physical robot, applying to other robotic tasks, and learning policies from different viewpoints. 
%%%%%%%
\begin{table}[t]
\caption{Success ratio, with standard error in brackets, of the models deployed to the moving object reaching task in the robot workspace with (Arch. full) and without (Arch. base) the dynamic filter. } 
\label{tab:moving}
\begin{center}
\footnotesize 
\setlength\tabcolsep{1 pt}
\renewcommand\theadset{\def\arraystretch{.5}}%
\begin{tabular}{lc  ccc ccccc}
\Xhline{1.0pt}
\multicolumn{2}{c}{\textbf{Model}}  & \multicolumn{8}{c}{\textbf{Object}}     \\
\cmidrule(lr){1-2}
\cmidrule(lr){3-10}
{\textit{Method}} &  {\textit{Arch.}} &   \textit{ball} & \textit{cup-H} & \textit{cup-J} & \textit{rubiks} & \textit{apple} & \textit{mug} & \textit{can} & \textit{cracker}  \\
\Xhline{1.0pt}       
\multirow{4}{*}{\thead{Scratch}}           &  base   &{\thead{28.4 \\ (5.0)}} &{\thead{87.7 \\ (3.7)}} &{\thead{77.8 \\ (4.6)}} &{\thead{92.6 \\ (2.9)}} &{\thead{58.0 \\ (5.5)}} &{\thead{64.2 \\ (5.3)}} &{\thead{85.2 \\ (3.9)}} &{\thead{63.0 \\ (5.4)}}    \\
&  full   &{\thead{29.6 \\ (5.1)}} &{\thead{75.3 \\ (4.8)}} &{\thead{58.0 \\ (5.5)}} &{\thead{42.0 \\ (5.5)}} &{\thead{71.6 \\ (5.0)}} &{\thead{55.6 \\ (5.5)}} &{\thead{84.0 \\ (4.1)}} &{\thead{98.8 \\ (1.2)}} \\
\midrule
\multirow{4}{*}{\thead{OHPL}}       & base &  {\thead{67.9 \\ (5.2)}} &{\thead{82.7 \\ (4.2)}} &{\thead{54.3 \\ (5.5)}} &{\thead{90.1 \\ (3.3)}} &{\thead{69.1 \\ (5.1)}} &{\thead{66.7 \\ (5.2)}} &{\thead{56.8 \\ (5.5)}} &{\thead{66.7 \\ (5.2)}}     \\
  & full   & {\thead{79.0 \\ (4.5)}} &{\thead{66.7 \\ (5.2)}} &{\thead{84.0 \\ (4.1)}} &{\thead{75.3 \\ (4.8)}} &{\thead{92.6 \\ (2.9)}} &{\thead{91.4 \\ (3.1)}} &{\thead{90.1 \\ (3.3)}} &{\thead{76.5 \\ (4.7)}}    \\
\Xhline{1.0pt}
\end{tabular}
\end{center}
\vspace{-9pt}
\end{table}

\section*{Acknowledgement}
This work is supported in part by the CHIST-ERA program through the project CORSMAL, under the UK EPSRC grant EP/S031715/1, and the UK EPSRC grant NCNR EP/R02572X/1.

\bibliographystyle{IEEEtran}
\bibliography{IEEEabrv,main}

%%%%%%%%%%%%%%%%%%%%%%%%%%%%%%%%%%%%%%%%%%%%%%%%%%%%%%%%%%%%%%%%%%%%%%%%%%%%%%%%

\end{document}